\documentclass[final,5p,times,twocolumn]{elsarticle}

\usepackage{amssymb}
\usepackage{amsmath}

\usepackage{xcolor}         
\definecolor{ssp-red}{HTML}{AE1324}
\definecolor{ssp-blue}{HTML}{09376F}

\newcommand{\abcmark}[1]{#1}

\usepackage{algorithmic}
\usepackage{algorithm}

\usepackage{orcidlink}
\usepackage{booktabs}       
\usepackage{amsfonts}       
\usepackage{multirow}
\usepackage{colortbl}

\usepackage{array}
\usepackage{textcomp}
\usepackage{stfloats}
\usepackage{verbatim}
\usepackage{graphicx}
\usepackage{microtype}      

\journal{Journal of \LaTeX}

\makeatletter
\def\ps@pprintTitle{%
  \let\@oddhead\@empty
  \let\@evenhead\@empty
  \def\@oddfoot{\hfil
    \scriptsize
    © 2026. This manuscript version is made available under the CC-BY-NC-ND 4.0 license 
        \url{https://creativecommons.org/licenses/by-nc-nd/4.0/} \\
    \hfil\thepage\hfil
  }%
  \let\@evenfoot\@oddfoot
}
\makeatother

\begin{document}

\begin{frontmatter}



\title{Semantic-decoupled Spatial Partition Guided Point-supervised Oriented Object Detection}

\author[ict,ucas]{Xinyuan Liu}
\ead{liuxinyuan21s@ict.ac.cn}

\author[hdu]{Hang Xu\texorpdfstring{\corref{cor1}}{}}
\ead{hxu@hdu.edu.cn}

\author[ict,ucas]{Zirui Chen}

\author[ict]{Yike Ma}

\author[hdu]{Chenggang Yan}

\author[ict]{Feng Dai}
\ead{fdai@ict.ac.cn}

\cortext[cor1]{Corresponding Author.}

\affiliation[ict]{
            organization={Institute of Computing Technology, Chinese Academy of Sciences},
            addressline={}, 
            city={Beijing},
            postcode={100190}, 
            state={},
            country={China}}

\affiliation[ucas]{
            organization={University of Chinese Academy of Sciences},
            addressline={}, 
            city={Beijing},
            postcode={100190}, 
            state={},
            country={China}}

\affiliation[hdu]{
            organization={Hangzhou Dianzi University},
            addressline={}, 
            city={Hangzhou},
            postcode={310018}, 
            state={},
            country={China}}

\begin{abstract}

Given its ability to reduce annotation costs, weakly supervised learning based on single-point annotations has emerged as a research focus in oriented object detection. Compared with the classical teacher-student paradigm, the simple model paradigm (e.g., PointOBB-v2) can substantially further reduce resources required for training while ensuring strong performance. The latter exhibits greater potential for low-cost training, yet such methods still face challenges of insufficient sample assignment and poor pseudo-label quality.
In this paper, we propose a training-efficient framework named SSP, which synergizes rule-driven prior injection and data-driven label purification. 
Specifically, SSP introduces two designs: 1) Pixel-level Spatial Partition-based Sample Assignment, which compactly estimates the upper and lower bounds of object scales and mines high-quality positive samples and hard negative samples through spatial partitioning of pixel maps. 2) Semantic Spatial Partition-based Box Extraction, which derives instances from spatial partitions modulated by semantic maps and converts them into pseudo-boxes for supervising detectors.
Experiments on DOTA-v1.0 and other datasets demonstrate SSP’s superiority: it achieves +6.73\% mAP improvement compared with the baseline, while requiring only 2 hours of training time and 6 GB of GPU memory. Furthermore, when SSP is integrated with stronger detector, the mAP can reach 50.81\%.
The code is available at \url{https://github.com/antxinyuan/ssp}.
\end{abstract}


\end{frontmatter}



\section{Introduction}
\label{sec:intro}

In recent years, technological advancements have driven exponential growth in the scale of accessible imagery, presenting both opportunities and challenges for automated scene understanding. As a fundamental task, oriented object detection aims to perceive the positions, scales, and orientations of various objects in scenes, with wide applications in aerial images \cite{gui2024remote}, panoramic images \cite{xu2022pandora}, scene text\cite{naiemi2022scene}, autonomous driving\cite{mao20233d}.
\cite{yang2023h2rbox,yu2023h2rboxv2,wang2024explicit}, with point supervision \cite{luo2024pointobb,yu2024point2rbox,ren2024pointobbv2} standing out for its unique balance of annotation efficiency and semantic richness. Unlike bounding box, point annotations reduce labeling effort significantly while retaining positional cues essential for object localization. This makes point supervision particularly suitable for remote sensing scenarios, where manual annotation of thousands of densely packed instances is impractical.

\begin{figure*}[!t]
\centering
\includegraphics[width=0.85\linewidth]{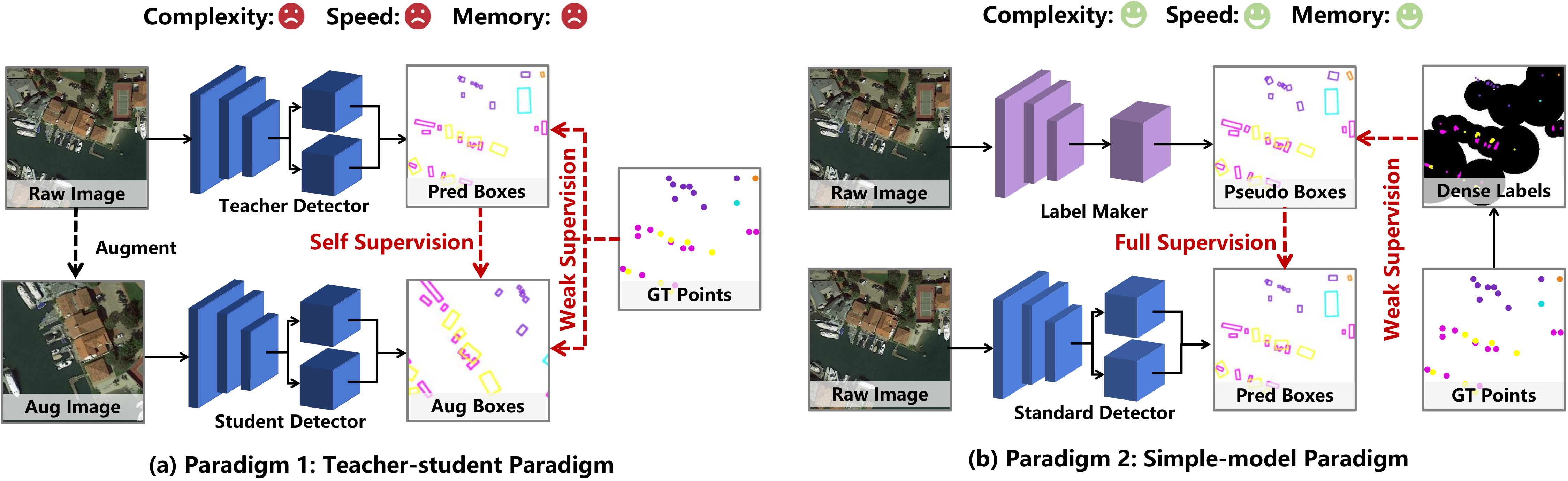}
\caption{Paradigm comparison of point-supervised oriented object detection. (a) Teacher-student paradigm \cite{luo2024pointobb,yu2024point2rbox,zhang2025pointobbv3} uses dual detectors for consistency-based supervision, with long training time and high resource consumption. (b) Simple-model paradigm \cite{ren2024pointobbv2,lu2025semantic} decouples pseudo label generation and detector training, significantly reducing training cost.}
\label{fig:paradigm}
\end{figure*}

\abcmark{Existing point-supervised oriented object detection methods can be divided into two dominant paradigms with different pipeline designs: the teacher-student paradigm \cite{luo2024pointobb,yu2024point2rbox,zhang2025pointobbv3,chen2022pointtobox} and the simple-model paradigm \cite{ren2024pointobbv2,lu2025semantic}. The \textbf{teacher-student paradigm}(Fig. \ref{fig:paradigm}(a)), the current mainstream approach inherited from horizontal bounding box supervision \cite{yang2023h2rbox}, adopts an online end-to-end joint training framework. A teacher model and student detector are optimized simultaneously, with missing scale and orientation information from point annotations mined from geometric consistency between the dual models' predictions. The \textbf{simple-model paradigm}(Fig. \ref{fig:paradigm}(b)), first introduced to oriented object detection by PointOBB-v2 \cite{ren2024pointobbv2}, follows an offline two-stage decoupled pipeline. A lightweight module is trained to convert point annotations into complete oriented pseudo bounding boxes. Then, the target detector is trained in a standard supervised manner with the pre-generated pseudo labels.}

\abcmark{The structural differences between the two paradigms directly lead to significant gaps in training efficiency and resource consumption. The teacher-student paradigm suffers from slow convergence and high overhead: its supervision signals must be mined gradually during end-to-end joint training, and dual-model co-optimization introduces persistent extra computational cost. In contrast, the simple-model paradigm decouples label generation and detector training, enabling an overall faster and more lightweight pipeline despite its two-stage structure: pseudo labels are generated via a lightweight module with minimal cost, and the second-stage detector converges much faster under explicit oriented box supervision.}
PointOBB-v2 \cite{ren2024pointobbv2}, as a representative work of this paradigm (distinguished from the PointOBB \cite{luo2024pointobb} and PointOBB-v3\cite{zhang2025pointobbv3}), introduces a simple-model framework that learns dense masks estimated from point annotations to generate pseudo bounding boxes for standard detector training. It offers two core advantages over the teacher-student paradigm: 1) \textbf{Higher training efficiency}: on the classic DOTA-v1.0 dataset, typical teacher-student models require over 10 hours of training and 10 GB of GPU memory, while PointOBB-v2 completes training within 1.5 hours with only 6 GB of memory, achieving remarkable acceleration; 2) \textbf{Higher label utilization}: unlike conventional methods that only treat fixed-radius regions around points as positive samples, it mines positive and negative samples via the spatial layout of point annotations, estimates object scale bounds, and introduces ignored samples to implicitly model potential object sizes. Its superior performance over complex teacher-student methods validates the promising potential of simple-model-based approaches.

\begin{figure}[!t]
\centering
\includegraphics[width=1.0\linewidth]{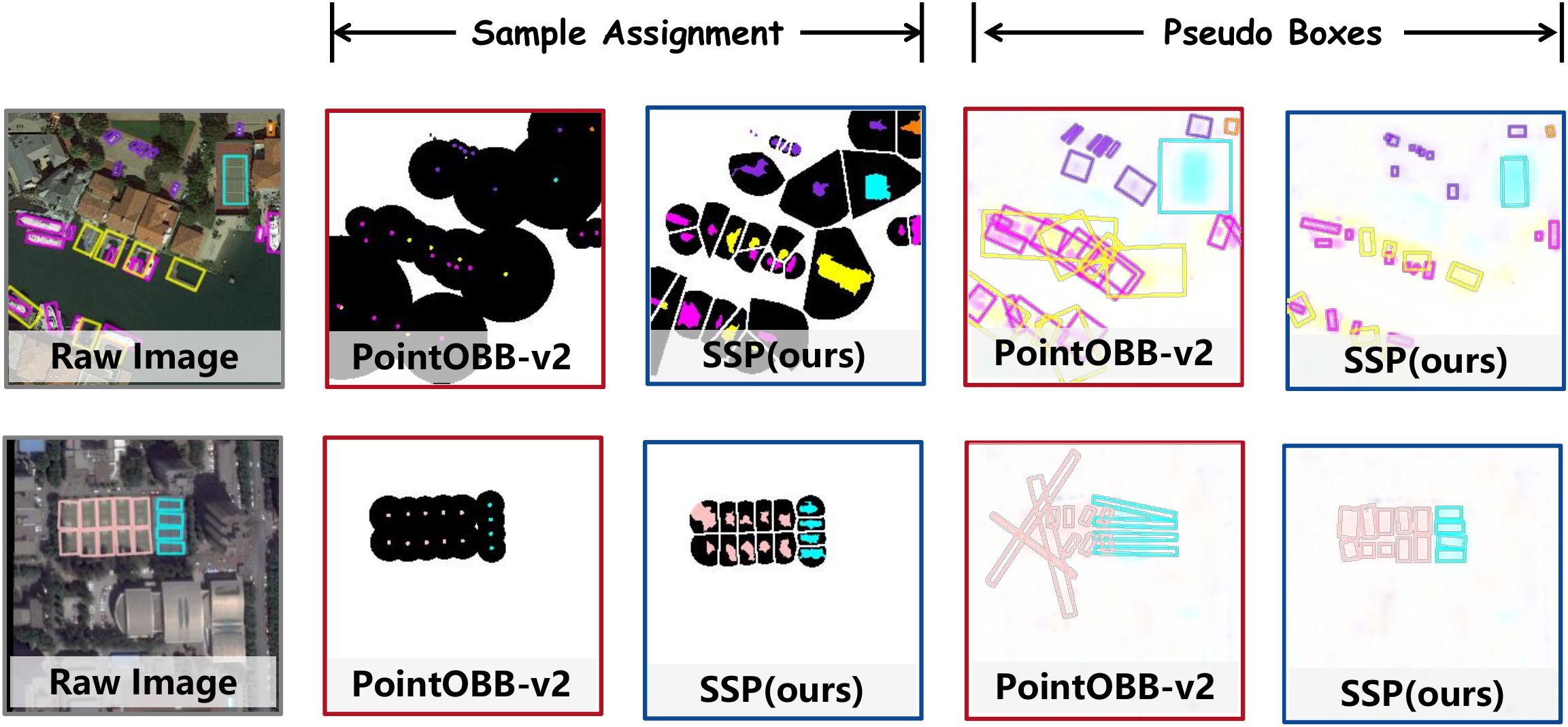}
\caption{\abcmark{Critical limitations in PointOBB-v2 \cite{ren2024pointobbv2}.} Here are two typical examples, where white represents the background, black denotes ignored regions, and colors indicate object boxes or masks. PointOBB-v2 suffers from inadequate sample assignment, characterized by excessively large ignored regions and extremely rare positive samples. After semantic learning, PointOBB-v2 exhibits unstable instance discrimination during box extraction. In contrast, our SSP enables more comprehensive sample assignment and more stable instance extraction of boxes.}
\label{fig:motivation}
\end{figure}

However, the current simple-model paradigm still has two critical non-negligible limitations, as shown in Fig. \ref{fig:motivation}:
\textbf{1) Inadequate sample assignment for mask learning.} Object scale estimation only provides upper bounds for defining ignore samples, while positive samples still rely on fixed-radius central regions; overlapping objects (e.g., ships \& harbors) also severely distort scale estimation, leading to serious underestimation.
\textbf{2) Unstable instance discrimination for box extraction.} Its point-oriented mask-to-box conversion is vulnerable to adjacent object interference in dense scenes, causing oversized bounding boxes, and it is highly sensitive to orientation estimation errors.
\abcmark{
Given that oriented object detection is predominantly applied to tasks with inherent top-down orthographic (bird's-eye view, BEV) imaging perspectives, these scenarios share two stable, well-recognized priors that directly mitigate the aforementioned limitations: first, objects exhibit minimal spatial occlusion in most scenes, eliminating the core interference for accurate object scale estimation; second, foreground objects maintain consistent appearance and texture under orthographic imaging, with no intra-class variations induced by viewpoint shifts, and present clearly distinguishable feature characteristics from the background. These priors lay a solid geometric and statistical foundation for our spatial partition-based pseudo label generation pipeline.}

Based on these characteristics, we propose a training-efficient point-supervised oriented object detection method guided by \textbf{S}emantic-decoupled \textbf{S}patial \textbf{P}artition (\textbf{SSP}). It inherits the advantage of high training efficiency from the PointOBB-v2\cite{ren2024pointobbv2} method, and significantly boosts detection performance through the more robust pseudo-label generation. 
\abcmark{The core philosophy of our method can be summarized as a sentence, i.e., introduce strong prior knowledge through rule-based sample assignment and then filter label noise via model learning to obtain reliable pseudo labels.}
Specifically, SSP includes two key designs with targeted core innovations:
\textbf{1) Pixel spatial partition-based sample assignment.} 
\abcmark{The core innovation is tight dual-bound estimation of object scales (both upper and lower bounds) via spatial partition and region growing, which breaks through the limitation of baseline methods that only provide rough upper-bound scale estimation.} Beyond dynamic radius-based masks, we generate a spatial partition map from annotated points, using dividing lines as additional hard negative samples to enhance instance discrimination, and partition-guided instance masks as additional positive samples to mine high-quality training samples.
\textbf{2) Semantic spatial partition-based box extraction.} 
\abcmark{The core innovation is upgrading classical spatial partition and region growing operations from raw pixel space to class-decoupled semantic feature space, which simultaneously suppresses low-level imaging noise and eliminates inter-class overlap interference.} We integrate class-decoupled semantic maps into the spatial partition framework, and derive instance masks from semantic maps rather than raw images. We further mitigate overlap interference via category-specific compatibility constraints, and convert refined instance masks into oriented boxes via the PCA-minmax method.
Finally, our main contribution can be summarized as follows:
\begin{enumerate}

\item We introduce a spatial partition-based sample assignment with region growing, which leverages the spatial arrangements and implicit appearance similarity in point annotations to compactly estimate object scale bounds and generate high-quality training samples.
\item We develop a learning-based pseudo-label purification mechanism that integrates spatial partition region growing with class-decoupled semantic maps, addressing instance omissions and errors inherent in early-stage sample assignment.
\item We present a training-efficient point-supervised oriented object detection framework under the simple-model paradigm, \abcmark{which fully retains the native training efficiency of this paradigm while significantly improving detection performance}, validated through extensive cross-dataset and cross-architecture experiments.

\end{enumerate}

The rest of this paper is organized as follows. Section \ref{sec:related_works} introduces related work of oriented object detection. Section \ref{sec:method} elaborates on our analysis and method. Section \ref{sec:exp} shows experimental results and comparison with other methods. Finally, conclusions are drawn in Section \ref{sec:conclusion}.

\section{Relation Works}
\label{sec:related_works}
\subsection{Fully Supervised Oriented Object Detection}

Oriented object detection (OOD) is well-suited for complex scenes \cite{naiemi2022scene,mao20233d,liu2023sph2pob,xu2023gldl}, especially in satellite remote sensing imagery where objects exhibit diverse orientations. This is attributed to its more accurate bounding box representations,i.e., oriented rectangles with explicit orientation angles.
Within the general object detection framework \cite{ren2015faster,tian2019fcos,carion2020end}, OOD has evolved specialized architectures. Two-stage detectors, such as Rotated cascade R-CNN \cite{zhu2019rotated} and Oriented R-CNN \cite{xie2021oriented}, refine region proposals into precise RBBs, excelling in high-precision scenarios. One-stage detectors like R$^3$Det \cite{yang2021r3det} and S$^2$A-Net \cite{han2022align} enable end-to-end dense prediction, optimizing inference efficiency. Transformer-based detectors, such as RoMP-transformer \cite{moon2024romp}, ARS-DETR \cite{zeng2024ars}, are designed to model long-range dependencies and multi-scale features.

Research on fully supervised Oriented Object Detection (OOD) primarily focuses on three challenges:
1) Bounding box representation: Methods like RepPoint-based approaches \cite{li2022oriented} represent objects via point sets, while Gaussian-based losses \cite{yang2021rethinking,xu2023rotated} model RBBs as probabilistic distributions to address parameter discontinuity.
2) Rotation-equivariant feature learning: Networks incorporating group-based modules \cite{ding2018learning,lee2024fred} and dynamic convolution modules \cite{huang2024discriminative,shi2025progressive} exhibit directional modeling capabilities in feature extraction.
3) Boundary discontinuity in angle regression: Smooth losses, e.g., SCRDet \cite{yang2019scrdet}, LinearGau \cite{zhou2024linear} alleviate this issue at the loss level, while angle coders (e.g., CSL \cite{yang2020arbitrary}, PSC \cite{yu2023psc}, ACM \cite{xu2024acm}) and angle-free representation (e.g., Oriented Reppoint \cite{li2022oriented}) further eradicate it at the model level.

As these challenges are addressed, fully supervised models have approached performance saturation: 85\% mAP has become a bottleneck on the DOTA benchmark, with few recent methods exceeding this threshold. However, high oriented bounding box annotation costs restrict their scalability to large-scale remote sensing datasets. Researchers are thus exploring low-cost alternatives like weakly/self-supervised learning to maintain performance with minimal labeling effort.

\subsection{Weakly Supervised Oriented Object Detection}
Weakly supervised oriented object detection focuses on leveraging weakened annotations to guide models in learning rotated bounding box (RBox) prediction. Due to significant reduction in annotation costs, it has become one of the most critical OOD research topics. According to the degree of label weakening, the research can be further subdivided into three categories: image-level, horizontal bounding box (HBox)-level, and point-level supervision.

\subsubsection{Image-supervised methods}
Image-level weak supervision provides labels that only provide the category of objects present in the entire image, representing the coarsest form of weak supervision. Although this setup has been extensively studied in general natural images \cite{zhang2021weakly,zhu2024misa}, it remains scarce in the remote sensing field \cite{tan2023wsodet}. In daily life scenes, a single image typically contains few objects, whereas remote sensing images often include numerous objects of the same type. This leads to highly homogeneous class labels across images, hindering models from mining effective information and distinguishing spatial distributions or orientation patterns within densely packed scenes.

\subsubsection{HBox-supervised methods}
Horizontal bounding box (HBox)-level weak supervision provides labels with instance-level HBoxes and categories, omitting only orientation information. Unique to oriented object detection, this setup balances information loss with reduced annotation costs, serving as a middle ground between fully supervised and image-level supervision. H2RBox \cite{yang2023h2rbox}, the seminal work under this setup, introduces a teacher-student framework. It utilizes geometric constraints to restrict object angles to discrete candidates, and refines predictions in a self-supervised branch, achieving robust orientation estimation. H2RBox-v2 \cite{yu2023h2rboxv2} improves upon this by leveraging object reflection symmetry to enhance RBox alignment with object extents. EIE-Det \cite{wang2024explicit} introduces explicit (rotation-equivariant) and implicit (scale/position-consistent) modules, enabling invariant feature learning across orientations, critical for scenes with diverse object distributions. Some studies~\cite{iqbal2021leveraging} use additional data for training, which are attractive but less general.

\begin{figure*}[!t]
\centering
\includegraphics[width=0.7\linewidth]{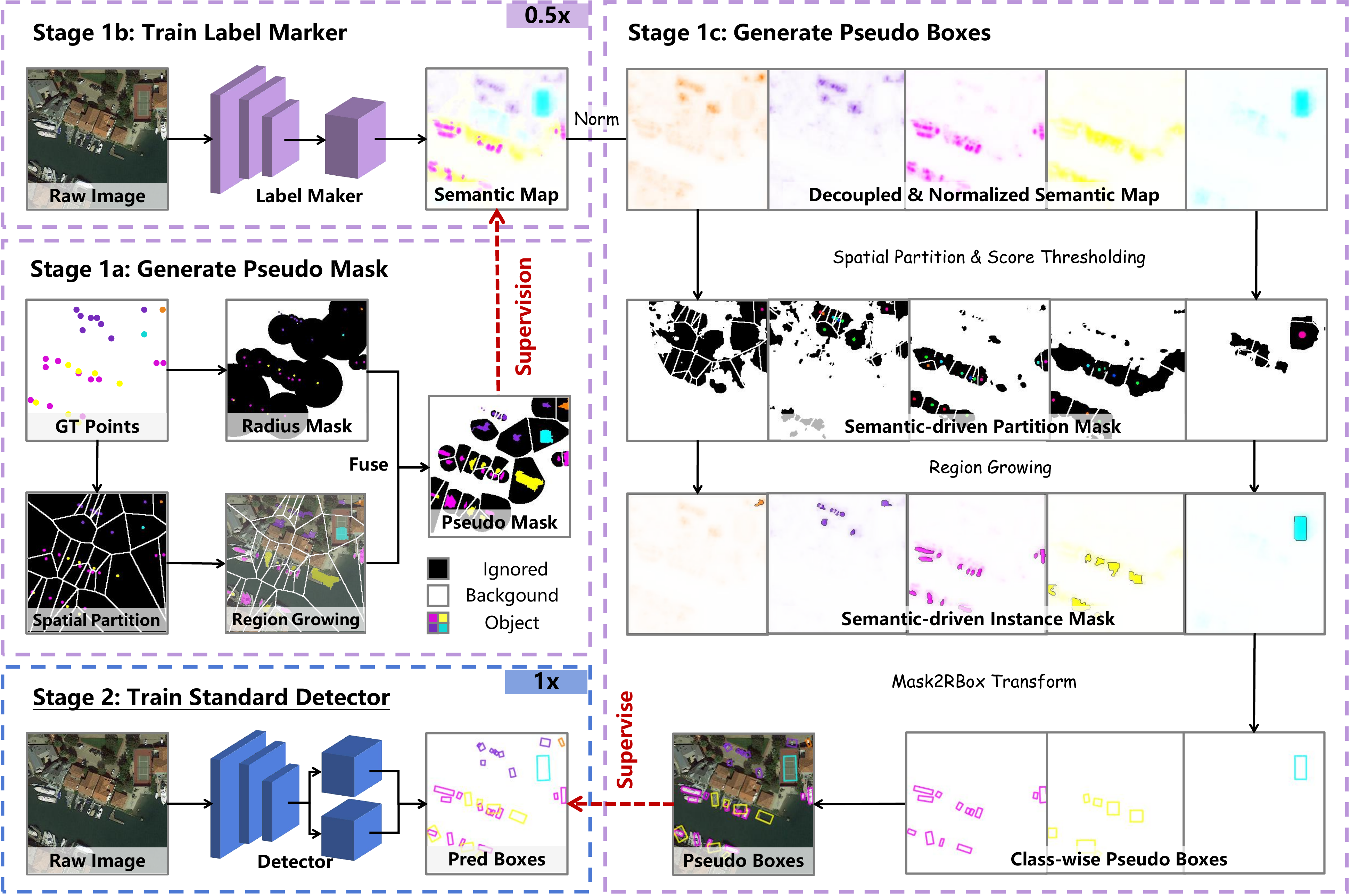}
\caption{\textbf{S}emantic-decoupled \textbf{S}patial \textbf{P}artition framework. The framework unfolds in two sequential stages, designed to leverage point supervision for robust oriented object detection.
In the stage 1, a pseudo mask is firstly generated from raw images with ground-truth (GT) points, via the fusion of dynamic radius-based assignment and spatial partition-based assignment. This fused pseudo mask serves as supervision for training the label marker, enabling it to learn class-awared spatial distributions. Subsequently, to generate pseudo boxes, instance masks are extracted from the decoupled semantic map, still relying on core operations: spatial partition to segment potential regions and region growing to refine object boundaries. Lastly, a PCA-minmax-based method is used to transform these masks into oriented bounding boxes. In the stage 2, an standard detector is trained using the generated pseudo labels.}
\label{fig:overview}
\end{figure*}

\subsubsection{Point-supervised methods}
Point-supervised labels provide only a single point and corresponding category for each instance, discarding not only orientation but also scale information. This setup further reduces annotation costs while maintaining instance-level supervision—critical for object-dense-packed scenes where it still conveys spatial object distribution. Following the teacher-student paradigm introduced by H2RBox \cite{yang2023h2rbox,yu2023h2rboxv2} and Point2RBox series \cite{yu2024point2rbox,yu2025point2rbox-v2} incorporate synthetic objects as pseudo-strong supervision to aid point-based learning, while PointOBB \cite{luo2024pointobb} and its journal-expanded version \cite{zhang2025pointobbv3} leverage scale consistency and multi-instance learning for additional supervision. However, teacher-student architectures inherently incur higher computational overhead.
Additionally, methods like PointSAM \cite{liu2025pointsam} and Point2Rbox-v3 \cite{zhang2025point2rbox-v3} exploit SAM’s \cite{kirillov2023segment} zero-shot capabilities for point supervision, but SAM’s reliance on massive pre-trained supervised data makes their classification as purely weakly-supervised debatable. Notably, the recent PointOBB-v2 \cite{ren2024pointobbv2} pioneers state-of-the-art performance in point-supervised oriented detection via a pseudo-labeling paradigm, explicitly eschewing traditional teacher-student architectures. It innovatively deciphers scale cues from multi-point layout and derives pseudo bounding boxes from class probability maps. The resulting pseudo labels are employed to train detectors, verifying the efficacy of pseudo-label learning frameworks.

Our approach follows the technical framework of PointOBB-v2 \cite{ren2024pointobbv2}. However, in sample assignment, we employ spatial partitioning to augment positive and negative samples. During the bounding box extraction stage, we discard the original point-axis search method and instead employed spatial partitioning with region growing to explicitly extract instance masks, which are then further converted into bounding boxes as pseudo-labels. When estimating object orientation, PointOBB-v2 requires probabilistic sampling within fixed regions due to its lack of instance shape awareness, whereas our method completely circumvents this issue.  
Through our dedicated efforts, the pseudo-labeling paradigm has achieved substantial advancements.

\section{Methods}
\label{sec:method}

\subsection{Overall Pipeline}
In the point-supervised training setting, only object center coordinates and class labels are available for each input image. As shown in Fig. \ref{fig:overview}, to deal with the scarcity of label information, we leverage the two-stage training pipeline of PointOBB-v2 \cite{ren2024pointobbv2}, where pseudo-labels are generated from the image content and point annotations at the $1^{st}$ stage. These pseudo-labels serve as comprehensive supervision for training a subsequent standard object detector at the $2^{nd}$ stage. The core contribution of this study lies in the analysis and design of the label marker in the first stage.

Following PointOBB-v2 \cite{ren2024pointobbv2}, the label marker in this work also employs an extremely simple architecture. Specifically, given an input image $I\in \mathbb{R}^{H \times W \times 3}$, semantic map (i.e., Class Probability Map (CPM) named in PointOBB-v2), $\bar{M} \in \mathbb{R}^{\bar{H} \times \bar{W} \times N_{cls}}$ can be obtained as follows: an image encoder extracts multi-scale features, where the highest-resolution feature map (i.e., P2) is then projected through a projection layer to generate the semantic map. The process can be formulated as
\begin{equation}
    \bar{M} = \textit{Proj}(f(I)[0])\abcmark{,}
\end{equation}
where $\textit{Proj}$ denotes the projection layer, consisting of four 256-channel convolution layers; $f$ represents a ResNet-50 \cite{he2016deep} backbone with FPN \cite{lin2017focal}.

To drive the training of the label marker, we employ the Pixel Spatial Partition Algorithm (details will be discussed in the Subsection \ref{sec:method_spsa}) to construct dense targets from sparse point annotations. The process can be formalized as
\begin{equation}
M = \textit{PSPSA}(I, S, G, C)\abcmark{,}
\end{equation}
where $\textit{PSPSA}$ represents \textit{Pixel Spatial Partition Sample Assignment} as defined in Algorithm \ref{alg:spsa}. $I$, $S$, $G$, $C$ denote raw image, sample points, ground-truth points, ground-truth classes, respectively. $M$ denotes assigned result, which can be also regarded as a pseudo mask.

By leveraging the above pseudo mask as the ground truth target, the label marker enables supervised learning, with FocalLoss \cite{lin2017focal} used as the loss function for pixel-wise classification during training:
\begin{equation}
\mathcal{L}_{pse} = \mathcal{L}_{cls}(\bar{M}, M)\abcmark{,}
\end{equation}
where $\bar{M}$, $M$ denote predicted semantic map and pseudo mask target, respectively. 

\begin{figure}[!t]
\centering
\includegraphics[width=\linewidth]{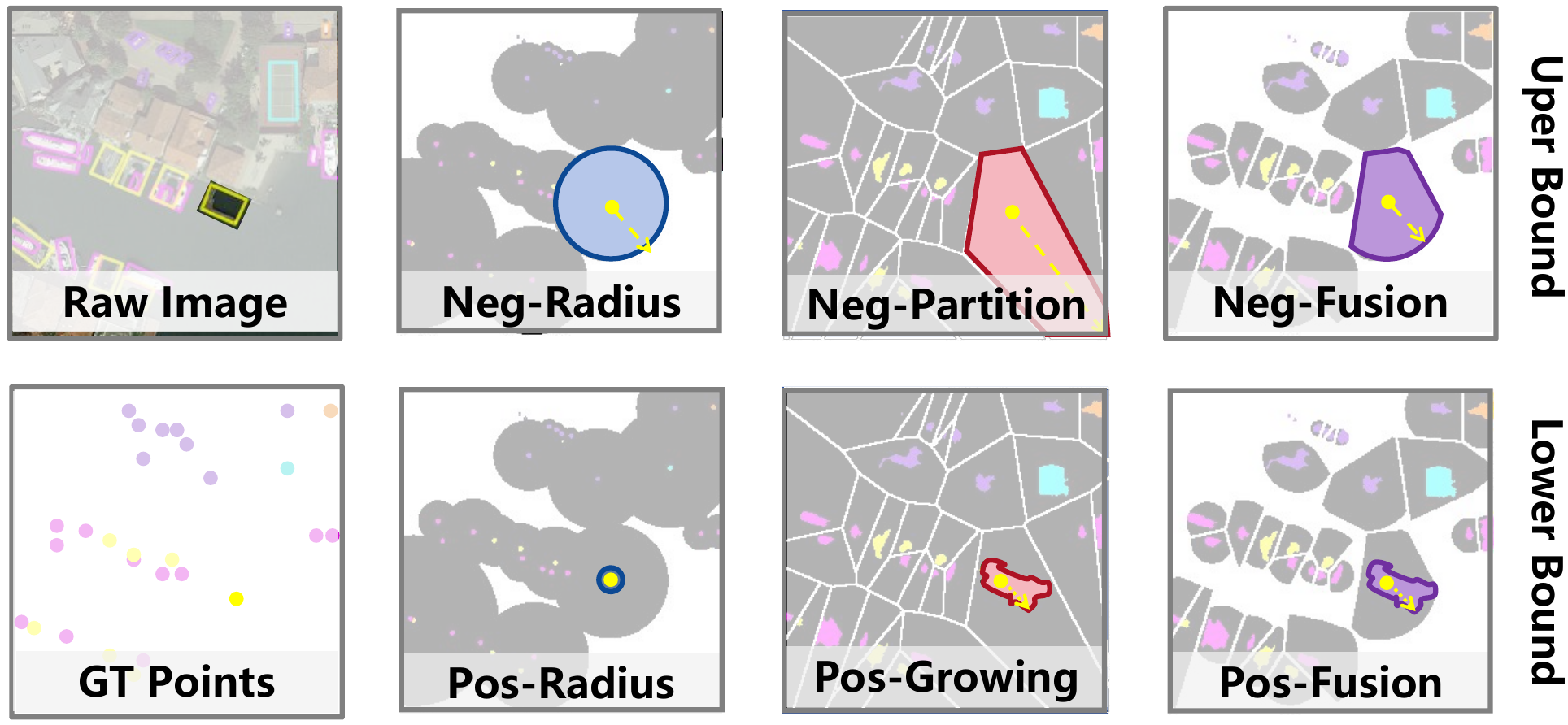}
\caption{\abcmark{Sample assignment fusion details.} For each instance ground-truth (e.g., the harbor marked by a yellow point), dynamic radius-based assignment computes circular upper and lower bounds (blue region), while partition\&growing-based assignment generates a polygonal upper bound and an irregular lower bound (red region).
\abcmark{The fused upper bound (purple region) is the spatial intersection (minimum value) of the two individual upper bounds, while the fused lower bound (purple region) is the spatial union (maximum value) of the two individual lower bounds, resulting in a tighter object scale range estimation.}}
\label{fig:assignment}
\end{figure}

\begin{algorithm}[t]
\caption{\textbf{P}ixel \textbf{S}patial \textbf{P}artitioning \textbf{S}ample \textbf{A}ssignment}
\label{alg:spsa}
\resizebox{0.85\linewidth}{!}{
\begin{minipage}{\linewidth}
\begin{algorithmic}[1]
\REQUIRE 
    Raw image $I \in \mathbb{R}^{H \times W \times 3}$, 
    GT points $G \in \mathbb{R}^{K \times 2}$, 
    GT classes $C \in \mathbb{Z}^{K \times 1}$
\ENSURE 
    Assignment result $M \in \mathbb{Z}^{N \times 1}$, where 
    $M[i] \in \{ign, bg, 1, ..., N_{cls}\}$

\STATE \textcolor{ssp-red}{\textbf{Step 1: Dynamic Radius-based Assignment}}
\FOR{$j \gets 1$ \TO $K$}
    \STATE $T[j] \gets \min_{i \neq j} \|G[j] - G[i]\|$ \textcolor{ssp-blue}{$\triangleright$ Estimate radius for each gt based on distance with nearest neighbor}
\ENDFOR

\FOR{$i \gets 1$ \TO $N$}
    \STATE $(\bar{d}, \bar{j}) \gets \arg\min_j \|S[i] - G[j]\|$
    \STATE $\tau^- \gets T[\bar{j}]$; $\tau^+ \gets$ fixed hyperparameter
    \IF{$\bar{d} < \tau^+$}
        \STATE $M[i] \gets C[\bar{j}]$ 
    \ELSIF{$\bar{d} > \tau^-$}
        \STATE $M[i] \gets bg$ 
    \ELSE
        \STATE $M[i] \gets ign$ 
    \ENDIF
    
\ENDFOR

\STATE \textcolor{ssp-red}{\textbf{Step 2: Spatial Partation-based Assignment}}
\STATE $P \gets \text{SpatialPartition}(G)$ 
\STATE $R \gets \text{RegionGrowing}(P, G, I)$
\STATE $V \gets \text{ValidateRegions}(R)$ \textcolor{ssp-blue}{$\triangleright$ Mark area-outlier regions}\strut

\FOR{$i \gets 1$ \TO $N$}
    \STATE $\bar{j} \gets \text{RegionOf}(R, S[i])$ \textcolor{ssp-blue}{$\triangleright$ Extract region of $i$-th sample}\strut
    \IF{$M[i] = ign$}
        \IF{$j \in [1, ..., N_{cls}]$ \AND $V[j]$}
            \STATE $M[i] \gets C[\bar{j}]$ 
        \ELSIF{$\text{IsBoundary}(S[i], P)$}
            \STATE $M[i] \gets bg$ 
        \ENDIF
    \ENDIF
\ENDFOR

\textbf{returns} $M$
\end{algorithmic}
\end{minipage}}
\end{algorithm}

After training, the label marker performs inference on the entire training set to generate dense semantic map. Compared with the rule-based masks constructed from sparse point annotations and raw image, model-generated ones exhibit stronger robustness and reliability. Critically, the generated maps provide full-image coverage, and even previously ignored regions without class assignments also receive appropriate class predictions, enabling easier perception of object shapes and scales.
On the generated semantic maps, we employ the Semantic Spatial Box Extraction Algorithm (details will be discussed in the Subsection \ref{sec:method_spbe}) to obtain pseudo labels. The process is formalized as
\begin{equation}
\mathcal{B}, \mathcal{C} = \textit{SSPBE}(\bar{M}, G, C)\abcmark{,}
\end{equation}
where $\textit{SSPBE}$ represents \textit{Semantic Spatial Partition Box Extraction} as defined in Algorithm \ref{alg:spbe}. $\bar{M}$, $G$, $C$ denote predicted semantic map, ground-truth points, ground-truth classes, respectively. $\mathcal{B}$, $\mathcal{C}$ denote extracted pseudo boxes and classes.

Finally, we utilize the pseudo-labels to train an additional standard detector, and the training loss function remains consistent with the conventional formulation:

\begin{equation}
\mathcal{L}_{det} = \mathcal{L}_{box}(\bar{B}, \mathcal{B}) + \mathcal{L}_{cls}(\bar{C}, \mathcal{C})\abcmark{,}
\end{equation}
where $\bar{B}$, $\bar{C}$ denote predicted boxes and classes, respectively; $\mathcal{B}, \mathcal{C}$ denote pseudo boxes and classes, respectively. It is worth noting that both the class loss $\mathcal{L}_{cls}$ and box loss $\mathcal{L}_{box}$ actually performed on dense prediction and targets, and the box loss $\mathcal{L}_{box}$ is only computed for positive samples. The formula here is a simplified expression for clarity.

\subsection{Pixel Spatial Partition based Sample Assignment}
\label{sec:method_spsa}

Sample assignment aims to surpass sparse point annotations by providing abundant training samples for the label marker. To this end, we designed the \textbf{P}ixel \textbf{S}patial \textbf{P}artitioning-based \textbf{S}ample \textbf{A}ssignment (PSPSA) as described in Algorithm \ref{alg:spsa}, which comprises two core steps: dynamic radius-based assignment and pixel spatial partition-based assignment. This design seeks to maximize the exploitation of supervisory information implied in images and annotations, as shown in Fig. \ref{fig:overview} and Fig. \ref{fig:assignment}.

\subsubsection{Dynamic radius-based Assignment}

\begin{figure}[!t]
\centering
\includegraphics[width=\linewidth]{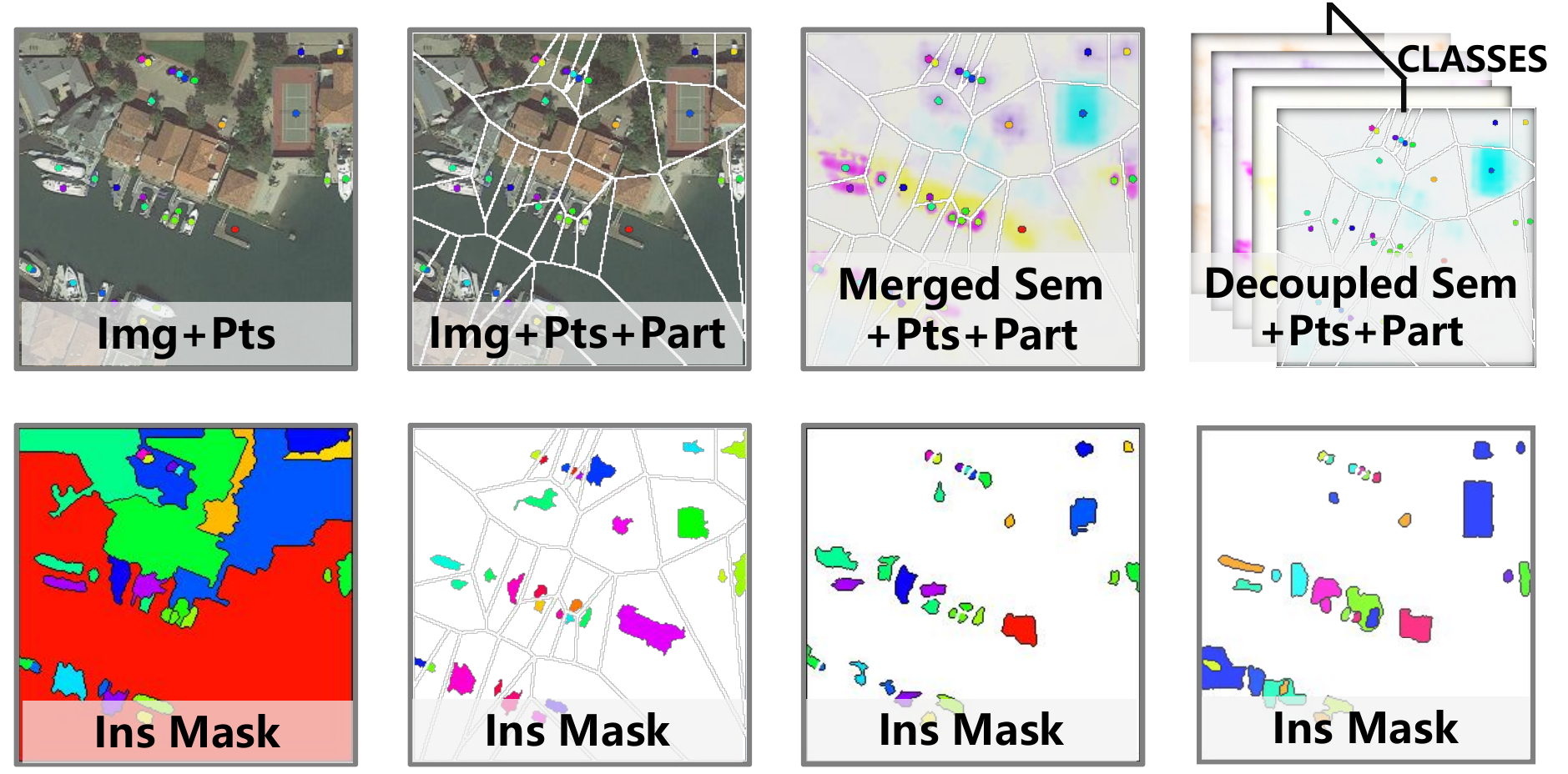}
\caption{\abcmark{Instance extraction details.} The first row depicts four settings, and the second row presents the corresponding \abcmark{results of} instance masks, where \textit{img} denotes raw image, \textit{pts} denotes ground-truth points, \textit{part} denotes spatial partition and 
\textit{ins} denotes instance. Without scale constraints from spatial partition boundaries, instance masks derived via region growing \abcmark{on the raw image} are prone to arbitrary over-expansion. With spatial partitioning, both raw image and merged semantic map-based instance extraction perform significantly better. Furthermore, \abcmark{class-decoupled} semantic map enables instance masks to achieve more accurate shapes and ensures overlapping object groups do not interfere with each other.}
\label{fig:instance}
\end{figure}

In dynamic radius-based sample assignment (Lines 2-16 in Algorithm \ref{alg:spsa}), the core task is to determine the lower and upper bounds of object scales. For the lower bound, zero represents the most conservative estimation. Fundamentally, we assume that a small region around the annotation point should belong to the object; thus, the lower bound is tightened from $0$ to a small positive value \abcmark{(default threshold: $\tau^+=5$)}. For the upper bound, we estimate it for each ground-truth (GT) point based on its distance to the nearest neighboring GT point. The value defines the maximum plausible size of the object, and the regions beyond this radius are confidently classified as background. This assumption is grounded in the fact that objects in remote sensing images rarely overlap under bird's-eye view. When the current object reaches its maximum size while the adjacent object becomes minimal, the adjacent object is reduced to a point, and the current object exactly envelopes it.

After defining the lower and upper bounds, we can obtain the initial sample assignment results. For each sample, if its distance to the nearest ground-truth (GT) point is smaller than the lower bound, it is marked as a positive sample; if the distance exceeds the upper bound, it is marked as a negative sample; otherwise, it is treated as an ignored sample. Ignored samples do not contribute to the model's loss computation, and their final class labels are determined by the trained model. The entire operation can be formalized as
\begin{equation}
M[i] = 
\begin{cases} 
C[\bar{j}] & \text{if } \bar{d} < \tau^+; \\
\text{bg} & \text{if } \bar{d} > \tau^-; \\
\text{ign} & \text{otherwise},
\end{cases}
\end{equation}
where $\bar{d}$ denotes the distance between $i$-th sample with the nearest $\bar{j}$-th gt; $\tau^+$, $\tau^-$ denote the lower \& upper scale bounds, respectively. $C[\bar{j}]$ denotes the class of $\bar{j}$-th gt; bg, ign denote background \& ignored sample.  
The above strategy is also mentioned in PointOBB-v2\cite{ren2024pointobbv2}, which is equivalent to only selecting high-quality samples for training and then using the model to generalize low-quality samples, thereby obtaining reliable mask predictions.

\subsubsection{Spatial partition-based Assignment}
The above strategy estimates the lower and upper bounds of scale via extreme cases, resulting in ignores many exploitable samples, whereas the bounds can actually be tightened further. To this end, we propose spatial partitioning-based assignment (Lines 17–30 in Algorithm \ref{alg:spsa}), whose core operations are spatial partitioning and region growing.

Spatial partitioning is used to further estimate the upper bound by assigning each sample in the space to the nearest gt instance, fully filling the space with these instance regions. This assumes that adjacent objects at most touch rather than contain each other, which \abcmark{better aligns with} most scenarios, not just edge cases. Intuitively, instance regions from spatial partitioning are generally mutually exclusive polygons, whereas those from dynamic radius are overlapping circles. The classic computational geometry Voronoi \cite{aurenhammer1991voronoi} method essentially performs similar operations and avoids regional fragmentation through global optimization. The entire operation can be formalized as
\begin{equation}
    P = \textit{SpatialPartition}(G)\abcmark{,}
\end{equation}
where $P$ denotes partition result, $G$ denote ground-truth points.

Region growing, meanwhile, is employed to refine the lower bound. It gradually expands regions from seed points (i.e., annotation points) using low-level image information, effectively augmenting positive samples. This approach is well-suited for remote sensing images, where objects under bird's-eye view typically exhibit simple visual appearances, consistent color and texture within objects, and distinct differences from the background. Additionally, spatial partitioning boundaries isolate multiple seed points during growth, preventing excessive region expansion, as shown in Fig. \ref{fig:instance}. The entire operation can be formalized as
\begin{equation}
    R = \textit{RegionGrowing}(P, G, I)\abcmark{,}
\end{equation}
where $R$ denotes growing result; $P$, $G$, $I$ denote partition result, ground-truth points and raw image, respectively.

Ultimately, both upper and lower scale bounds are tightened, upgraded from isotropic to anisotropic, and better aligned with the requirements of rotated object detection tasks. The new operation can be formalized as
\begin{equation}
M[i] = 
\begin{cases} 
C[\bar{j}] & \text{if } \bar{d} < \bar{\tau}^{+}(\theta)\abcmark{;} \\
\text{bg} & \text{if } \bar{d} > \bar{\tau}^{-}(\theta)\abcmark{;} \\
\text{ign} & \text{otherwise}\abcmark{,}
\end{cases}
\end{equation}
where $\theta$ denotes the oriented angle of the $i$-th sample relative to the $\bar{j}$-th gt point; $\bar{\tau}^+(\theta)$, $\bar{\tau}^-(\theta)$ represent the $\theta$-orient distance between gt point and its and lower boundary curve (region growing outline) and upper boundary curve (spatial partition boundary), respectively.

\subsubsection{Assignment results fusion}
Further, we integrate the two strategies for sample assignment to complement each other, as shown in Fig. \ref{fig:assignment}. For upper bounds, we take the smaller value, as $\bar{\tau}^-(\theta)$ may unreliably extend to spatial boundaries without nearby objects in the $\theta$-direction. For lower bounds, we adopt the larger value. Considering similar scales of same-class objects, we filter obviously abnormal regions by growth area, reverting the lower bound to a fixed radius $\tau^+$ in such cases. The final operation can be formalized as
\begin{equation}
M[i] = 
\begin{cases} 
C[\bar{j}] & \text{if } \bar{d} < max(\tau^+, \bar{\tau}^{+}(\theta))\abcmark{;} \\
\text{bg} & \text{if } \bar{d} > min(\tau^-, \bar{\tau}^{-}(\theta))\abcmark{.} \\
\text{ign} & \text{otherwise}
\end{cases}
\end{equation}

\subsection{Semantic Spatial Partition based Box Extraction}
\label{sec:method_spbe}

\begin{algorithm}[t]
\caption{\textbf{S}emantic \textbf{S}patial \textbf{P}artitioning \textbf{B}ox \textbf{E}xtraction}
\label{alg:spbe}
\resizebox{0.85\linewidth}{!}{
\begin{minipage}{\linewidth}
\begin{algorithmic}[1]
\REQUIRE 
    Semantic map $\bar{M} \in \mathbb{R}^{\bar{H} \times \bar{W} \times N_{cls}}$,
    Sample points $S \in \mathbb{R}^{N \times 2}$, \\
    GT points $G \in \mathbb{R}^{K \times 2}$,
    GT classes $C \in \mathbb{Z}^{K \times 1}$
\ENSURE 
    Pseudo boxes $\mathcal{B} \in \mathbb{R}^{K \times 5}$, 
    Pseudo classes $\mathcal{C} \in \mathbb{R}^{K \times 1}$

\STATE Initialize $\mathcal{B} \gets \varnothing$; $\mathcal{C} \gets \varnothing$
\FOR{$c \gets 1$ \TO $N_{cls}$}
    \STATE \textcolor{ssp-red}{\textbf{Step 1: Class-wise Spatial Partitioning}}
    \STATE $\bar{M}_c \gets \text{Norm}(\bar{M}[c])$ \textcolor{ssp-blue}{$\triangleright$ Normalize class-$c$ semantic map}\strut
    \STATE $G_c \gets$ Select GT points belong to class-$c$
    \STATE $G_{cc} \gets$ Select GT points belong to compatible class-$cc$ 
    
    \STATE $P_c \gets \text{SpatialPartition}(G_c \cup G_{cc})$
    \STATE $P_s \gets \text{ScoreThresholding}(F_c, \tau)$ \textcolor{ssp-blue}{$\triangleright$ Seed preassignment}\strut
    \STATE $R_c \gets \text{RegionGrowing}(P_c \cap P_s, G_c, \bar{M}_c)$
    
    \STATE \textcolor{ssp-red}{\textbf{Step 2: Instance Box Conversion}}
    \FOR{$k \gets 1$ \TO $|G_c|$}
        \STATE $\mathcal{M} \gets \text{InstanceOf}(R_c, k)$ \textcolor{ssp-blue}{$\triangleright$ Extract $k$-th instance mask}\strut
        \STATE $b \gets \text{Mask2RBox}(\mathcal{M}, G_c[k])$
        \STATE $\mathcal{B} \gets \mathcal{B} \cup \{b\}$; $\mathcal{C} \gets \mathcal{C} \cup \{c\}$
    \ENDFOR
\ENDFOR

\RETURN $\mathcal{B}$, $\mathcal{C}$
\end{algorithmic}
\end{minipage}}
\end{algorithm}

Bounding box extraction aims to separate individual instances from dense masks, thereby upgrading point annotations to complete box annotations and supervising detector learning. To this end, we designed the \textbf{S}emantic \textbf{S}patial \textbf{P}artitioning-based \textbf{B}ox \textbf{E}xtraction (SSPBE) as described in Algorithm \ref{alg:spbe}, which consists of two core steps: class-wise spatial partitioning and instance box conversion. This design aims to fully leverage the potential of semantic mask learning, as shown in Fig. \ref{fig:instance}.

\subsubsection{Class-wise Spatial Partitioning}
Class-wise spatial partitioning (Line 2–10 in Algorithm \ref{alg:spbe}), a pivotal step  in pseudo-label generation, aims to separate individual object masks from semantic maps. Here, we reuse the pipeline of spatial partitioning with region growing, but apply it to decoupled semantic maps, as shown in Fig. \ref{fig:overview}.

Firstly, the predicted semantic map is class-wise splited and normalized to restore the predictive confidence of classes whose absolute numeric are suppressed by inter-class interference, thereby reducing interference during region growing, especially for nested category combination (e.g., \textit{harbor$\&$ship} or \textit{ground-track-field$\&$soccer-ball-field} in the DOTA dataset). Secondly, for each category, we select objects of the current class and its compatible categories to perform spatial partitioning as
\begin{equation}
    P_c = \textit{SpatialPartition}(G_c \cup G_{cc})\abcmark{,}
\end{equation}
where $G_c$, $G_{cc}$ denote ground-truth points of class-$c$ and its compatible class-$cc$. This approach still fully leverages the non-overlapping characteristic of most objects (even across different classes), while mitigating interference from a few incompatible (i.e., nested layout) category groups. 

Additionally, probabilistic scores in semantic maps serve as a filtering criterion to eliminate low-confidence regions \abcmark{(default threshold: $\tau=0.3$)} during instance extraction. Finally, we perform region growing on the current class's semantic map to obtain instances mask for each category, as shown in Fig. \ref{fig:instance} and following:
\begin{equation}
    R_c = \textit{RegionGrowing}(P_c \cap P_s, G_c, \bar{M}_c)\abcmark{,}
\end{equation}
where $P_c$, $P_s$ denote partition result and score-based filter result, respectively; $G_c$, $M_c$ denote ground-truth points and semantic map of class-$c$, respectively.

\begin{algorithm}[t]
\caption{Mask2Rbox (\textit{pca-minmax})}
\label{alg:mask2rbox-pca}
\resizebox{0.85\linewidth}{!}{
\begin{minipage}{\linewidth}
\begin{algorithmic}[1]
\REQUIRE 
    Instance mask points $\mathcal{M} \in \mathbb{R}^{N \times 2}$, 
    GT points $\boldsymbol{g} \in \mathbb{R}^{2}$,
\ENSURE 
    Rotated bounding box $\boldsymbol{b} \in \mathbb{R}^{5}$ 

\STATE $U, \Sigma, V \gets \text{PCA}(\mathcal{M})$ 
\STATE $\theta \gets \arctan({V_{1,0}}/{V_{0,0}})$ \textcolor{ssp-blue}{$\triangleright$ Compute rotation angle}

\STATE $\boldsymbol{c} \gets \text{Mean}(\mathcal{M})$  if $\boldsymbol{g}$ is \textit{none} \textbf{else} $\boldsymbol{g}$ 

\STATE $\mathcal{P}' \gets (\mathcal{P} - \mathbf{c}) \cdot V$ \textcolor{ssp-blue}{$\triangleright$ Project points to PCA basis}
\STATE $l, d \gets \text{Min}(\mathcal{P}')$ 
; $r, t \gets \text{Max}(\mathcal{P}')$ 

\STATE $w \gets 2 \cdot \max(|l|, |r|)$ 
; $h \gets 2 \cdot \max(|t|, |d|)$

\STATE $\boldsymbol{b} \gets (\boldsymbol{c}_x, \boldsymbol{c}_y, w, h, \theta)$

\RETURN $\boldsymbol{b}$
\end{algorithmic}
\end{minipage}}
\end{algorithm}

\subsubsection{Instance box conversion}

Instance box conversion (Line 11-18 in Algorithm \ref{alg:spbe}) is the last step of pseudo-label generation, converting instances from mask to rotated bounding boxes. Conventionally, numerous prior works \cite{li2022oriented,liu2025pointsam} have employed the off-the-shelf \texttt{minAreaRect()} method in OpenCV toolbox. This method utilizes the Rotating Calipers algorithm \cite{preparata2012computational} to compute the minimum-area bounding rectangle for a point set by enumerating edges of its convex hull. However, we discern that for several man-made objects (e.g., airplane \& helicopter), the desired orientation of the rotated rectangle typically aligns with the object’s symmetry axis, but such a criterion frequently unfulfilled by \texttt{minAreaRect()}.

To address this limitation, we introduce a new Mask2Rbox method named PCA-MinMax, as delineated in Algorithm \ref{alg:mask2rbox-pca}. For each instance mask, our approach first derives the principal direction via Principal Component Analysis (PCA). Subsequently, we compute the minimum and maximum coordinates of points along this principal direction and its orthogonal axis to construct the bounding box. By incorporating the spatial distribution of points, this strategy tends to ensure that the resulting box orientation closely matches the object’s symmetry axis.

\section{Experiments}
\label{sec:exp}

\subsection{Datasets}
We explored diverse remote sensing image datasets in our extensive experiments, including DOTA-v1.0/v1.5/v2.0 \cite{xia2018dota} and RSAR \cite{zhang2025rsar}.

\textbf{DOTA} \cite{xia2018dota} is one of the most popular datasets for oriented object detection in aerial images, comprising RGB and grayscale images, with the former sourced from Google Earth and CycloMedia, and the latter derived from the panchromatic bands of GF-2 and JL-1 satellite imagery. Collected across diverse sensors and platforms, the dataset’s images span sizes from 800×800 to 20,000×20,000 pixels, depicting objects with wide-ranging scales, orientations, and shapes. It currently includes three versions:
\begin{itemize}
    \item \textbf{DOTA-v1.0} includes 2,806 images, and 188,282 instances, with data split into training (1/2), validation (1/6), and testing (1/3) sets. It contains 15 classes: \textit{Plane (PL)}, \textit{Baseball diamond (BD)}, \textit{Bridge (BR)}, \textit{Ground track field (GTF)}, \textit{Small vehicle (SV)}, \textit{Large vehicle (LV)}, \textit{Ship (SH)}, \textit{Tennis court (TC), Basketball court (BC)}, \textit{Storage tank (ST)}, \textit{Soccer-ball field (SBF)}, \textit{Roundabout(RA)}, \textit{Harbor (HA)}, \textit{Swimming pool (SP)} and \textit{Helicopter (HC)}.
    \item \textbf{DOTA-v1.5} retains the same image pool but adds annotations for extremely small instances ($\leqslant$10 pixels) and introduces a new category, \textit{container crane (CC)}, totaling 403,318 instances.
    \item \textbf{DOTA-v2.0} further expands the dataset with additional Google Earth, GF-2 satellite, and aerial images, featuring 18 categories—including newly added \textit{airport (AP)} and \textit{helipad (HP)}—across 11,268 images and 1,793,658 instances. The images are partitioned into training (1,830 images, 268,627 instances), validation (593 images, 81,048 instances), test-dev (2,792 images, 353,346 instances). 
\end{itemize}

\textbf{RSAR} \cite{zhang2025rsar} is the largest multi-class rotated object detection dataset for Synthetic Aperture Radar (SAR) imagery to date. RSAR comprises 95,842 images, including 78,837 in the training set, 8,467 in the validation set, and 8,538 in the test set. It covers six typical classes: \textit{Ship (SH)}, \textit{Aircraft (AI)}, \textit{Car (CA)}, \textit{Tank (TA)}, \textit{Bridge (BR)}, and \textit{Harbor (HA)}. Officially, all images are cropped to 800×800 pixels.

\subsection{Implementation Details}
All experiments are conducted on NVIDIA RTX 3090 GPUs using PyTorch 1.10.0 and the rotation detection toolkit MMRotate 0.3.4 \cite{zhou2022mmrotate}. To ensure fairness, training configurations for all experiments generally follow the baseline PointOBB-v2 and other related works. The detector is implemented using the standard ResNet50-FPN-FCOS framework, trained with a standard 1× schedule (12 epochs) on all datasets. The label marker adopts the same architectural backbone as the detector but removes the box branch; its training schedule is halved to 0.5× epochs for all datasets. For both the detector and the label marker, an SGD optimizer is employed, initialized at 1e-2 for a batch size of 16, with a 500-iteration warm-up period and learning rate decay by a factor of 10 at each step. Data augmentation only employs random flipping. All experiments are evaluated without multi-scale technique \cite{zhou2022mmrotate}, and Average Precision (AP) is adopted as the primary metric. 
\abcmark{Regarding class compatibility matrix, any two classes are regarded as compatible by default. In the DOTA dataset series, exceptions only apply to four pairs: \textit{ship}\&\textit{harbor}, \textit{ground-track-field}\&\textit{soccer-ball-field}, \textit{plane}\&\textit{airport}, and \textit{helicopter}\&\textit{helipad}. In the RASR dataset, the only exception is the pair \textit{ship}\&\textit{harbor}. These settings are based on the observation that objects in these pairs frequently exhibit nested or heavily overlapped layouts in remote sensing scenes.}

\begin{table*}[t]
\centering
\resizebox{\linewidth}{!}{
\begin{tabular}{llcccccccccccccccc}
\toprule
\textbf{Methods} & \textbf{Epochs$^1$} & \textbf{PL}$^2$    & \textbf{BD}    & \textbf{BR}    & \textbf{GTF}   & \textbf{SV}    & \textbf{LV}    & \textbf{SH}    & \textbf{TC}    & \textbf{BC}    & \textbf{ST}    & \textbf{SBF}   & \textbf{RA}    & \textbf{HA}    & \textbf{SP}    & \textbf{HC}    & \textbf{mAP}  \\ \hline
\multicolumn{18}{l}{$\blacktriangledown$ \textit{RBox-supervised OOD}} \\ \hline
RetinaNet{\tiny ICCV'17} \cite{lin2017focal} & $12$& 88.2  & 77.0  & 45.0  & 69.4  & 71.5  & 59.0  & 74.5  & 90.8  & 84.9  & 79.3  & 57.3 & 64.7  & 62.7  & 66.5  & 39.6  & 68.69 \\
GWD{\tiny ICML'21}\cite{yang2021rethinking} & $12$ & 89.3  & 75.4  & 47.8  & 61.9  & 79.5  & 73.8  & 86.1  & 90.9  & 84.5  & 79.4  & 55.9 & 59.7  & 63.2  & 71.0  & 45.4  & 71.66 \\
FCOS{\tiny ICCV'19} \cite{tian2019fcos} & $12$ & 89.1  & 76.9  & 50.1  & 63.2  & 79.8  & 79.8  & 87.1  & 90.4  & 80.8  & 84.6  & 59.7 & 66.3  & 65.8  & 71.3  & 41.7  & 72.44 \\
S$^2$A-Net{\tiny AAAI'22} \cite{han2022align} & $12$ & 89.2  & 83.0  & 52.5  & 74.6  & 78.8  & 79.2  & 87.5  & 90.9  & 84.9  & 84.8  & 61.9 & 68.0  & 70.7  & 71.4  & 59.8  & \textbf{75.81} \\ \hline
\multicolumn{18}{l}{$\blacktriangledown$ \textit{HBox-supervised OOD}} \\ \hline
H2RBox{\tiny ICLR'23} \cite{yang2023h2rbox} & $12^*$ & 88.5  & 73.5  & 40.8  & 56.9  & 77.5  & 65.4  & 77.9  & 90.9  & 83.2  & 85.3  & 55.3 & 62.9  & 52.4  & 63.6  & 43.3  & 67.82 \\
EIE-Det{\tiny TETCI'24}\cite{wang2024explicit}  & $12^*$ & 87.7 & 70.2 & 41.5 & 60.5 & 80.7 & 76.3 & 86.3 & 90.9 & 82.6 & 84.7 & 53.1 & 64.5 & 58.1 & 70.4 & 43.8 & \underline{70.10} \\
H2RBox-v2{\tiny NIPS'23} \cite{yu2023h2rboxv2} & $12^*$ & 89.0 & 74.4 & 50.0 & 60.5 & 79.8 & 75.3 & 86.9 & 90.9 & 85.1 & 85.0 & 59.2 & 63.2 & 65.2 & 70.5 & 49.7 & \textbf{72.31} \\ \hline
\multicolumn{18}{l}{$\blacktriangledown$ \textit{Point-supervised OOD} (\textit{Teacher-student Model})} \\ \hline
\scriptsize P2BNet+H2RBox{\tiny ICLR'23} \cite{chen2022pointtobox,yang2023h2rbox} & $12^*+12$ & 24.7 & 35.9 & 7.1 & 27.9 & 3.3 & 12.1 & 17.5 & 17.5 & 0.8 & 34.0 & 6.3 & 49.6 & 11.6 & 27.2 & 18.8 & 19.63 \\
\scriptsize P2BNet+H2RBox-v2{\tiny NIPS'23} \cite{chen2022pointtobox,yu2023h2rboxv2} & $12^*+12$ & 11.0 & 44.8 & 14.9 & 15.4 & 36.8 & 16.7 & 27.8 & 12.1 & 1.8 & 31.2 & 3.4 & 50.6 & 12.6 & 36.7 & 12.5 & 21.87\\
\footnotesize Point2RBox{\tiny CVPR'24} \cite{yu2024point2rbox} & $12*$ & 62.9 & 64.3 & 14.4 & 35.0 & 28.2 & 38.9 & 33.3 & 25.2 & 2.2  & 44.5 & 3.4  & 48.1 & 25.9 & 45.0 & 22.6 & 34.07 \\
\footnotesize Point2RBox+SK$^\dagger${\tiny CVPR'24} \cite{yu2024point2rbox} & $12^*$ & 53.3 & 63.9 & 3.7  & 50.9 & 40.0 & 39.2 & 45.7 & 76.7 & 10.5 & 56.1 & 5.4  & 49.5 & 24.2 & 51.2 & 33.8 & 40.27 \\
PointOBB{\tiny CVPR'24} \cite{luo2024pointobb} & $24^*+12$ & 26.1 & 65.7 & 9.1 & 59.4 & 65.8 & 34.9 & 29.8 & 0.5 & 2.3 & 16.7 & 0.6 & 49.0 & 21.8 & 41.0 & 36.7  & 30.08 \\ 
PointOBB-v3{\tiny IJCV'25} \cite{zhang2025pointobbv3} & $24^*$ & 30.9 & 39.4 & 13.5 & 22.7 & 61.2 & 7.0 & 43.1 & 62.4 & 59.8 & 47.3 & 2.7 & 45.1 & 16.8 & 55.2 & 11.4 & 41.29 \\
PointOBB-v3{\tiny IJCV'25} \cite{zhang2025pointobbv3} & $24^*+12$ & 52.9 & 54.4 & 21.3 & 52.7 & 65.6 & 44.9 & 67.8 & 87.2 & 26.7 & 73.4 & 32.6 & 53.3 & 39.0 & 56.4 & 10.2 & 49.24 \\
Point2RBox-v2{\tiny CVPR'25}\cite{yu2025point2rbox-v2} & $12^*$ & 78.4 & 52.7 & 8.3 & 40.9 & 71.0 & 60.5 & 74.7 & 88.7 & 65.5 & 72.1 & 24.4 & 26.1 & 30.1 & 50.7 & 21.0 & \underline{51.00} \\
Point2RBox-v2{\tiny CVPR'25} \cite{yu2025point2rbox-v2} & $12^*+12$ & 88.0 & 72.6 & 8.0 & 46.2 & 79.6 & 76.3 & 86.9 & 89.1 & 79.7 & 82.9 & 26.2 & 45.3 & 45.8 & 66.3 & 46.3 & \textbf{62.61} \\ \hline
\multicolumn{18}{l}{$\blacktriangledown$ \textit{Point-supervised OOD} (\textit{Simple Model})} \\ \hline
Point2Mask-RBox{\tiny CVPR'23} \cite{li2023point2mask}  & $12$ & 4.0 & 23.1 & 3.8 & 1.3 & 15.1 & 1.0 & 3.3 & 19.0 & 1.0 & 29.1 & 0.0 & 9.5 & 7.4 & 21.1 & 7.1 & 9.72 \\
PointOBB-v2{\tiny ICLR'25} \cite{ren2024pointobbv2} & $6+12$ & 64.5 & 27.8 & 1.9 & 36.2 & 58.8 & 47.2 & 53.4 & 90.5 & 62.2 & 45.3 & 12.1 & 41.7 & 8.1 & 43.7 & 32.0 & 41.68 \\ 
PMS-SAM-RSD$^\dagger${\tiny EL'25} \cite{lu2025semantic} & 12 & 69.0 & 39.5 & 6.7 & 44.8 & 64.7 & 71.9 & 79.6 & 79.8 & 2.7 & 60.0 & 12.1 & 32.6 & 39.6 & 44.8 & 42.5 & \underline{46.00} \\
\rowcolor{gray!20} 
SSP(FCOSR){\tiny Ours} & $6+12$ & 78.1 & 51.6 & 0.0 & 42.0 & 51.2 & 26.9 & 54.5 & 90.0 & 68.3 & 73.2 & 33.3 & 35.8 & 30.4 & 55.9 & 34.3 & \textbf{48.41} \\
\rowcolor{gray!20} 
SSP(ORCNN){\tiny Ours} & $6+12$ & 80.1 & 55.3 & 0.3 & 47.7 & 51.0 & 32.8 & 53.8 & 90.4 & 73.6 & 74.8 & 38.8 & 26.8 & 32.5 & 53.1 & 38.9 & \textbf{50.00}\\
\rowcolor{gray!20} 
SSP(ReDet){\tiny Ours} & $6+12$ & 80.3 & 58.3 & 0.3 & 45.9 & 48.1 & 30.6 & 54.0 & 90.3 & 76.3 & 75.2 & 41.6 & 26.9 & 40.0 & 56.8 & 37.7 & \textbf{50.81}\\
\bottomrule
\specialrule{0pt}{2pt}{0pt}
\multicolumn{18}{l}{$^\dagger$Using strong priors. 
Point2RBox+SK: One-shot sketches for each class; PMS-SAM-RSD: SAM.} \\
\multicolumn{18}{l}{$^1$$^*$indicates data passes through teacher\&student two models  thereby doubling the equivalent epochs; $+$ denotes two-stage training.} \\
\multicolumn{18}{l}{$+$ denotes two-stage training under the pseudo-label paradigm.} \\
\multicolumn{18}{l}{$^2$PL: Plane, BD: Baseball diamond, BR: Bridge, GTF: Ground track field, SV: Small vehicle, LV: Large vehicle, SH: Ship, TC: Tennis court,} \\
\multicolumn{18}{l}{$\,\;$BC: Basketball court, ST: Storage tank, SBF: Soccer-ball field, RA: Roundabout, HA: Harbor, SP: Swimming pool, HC: Helicopter.} \\
\bottomrule
\end{tabular}}
\caption{Detection performance of each category on the DOTA-v1.0 and the mean mAP of all categories.}
\label{tab:exp_dota}
\end{table*}

\begin{table*}[!tb]
\centering
\resizebox{0.8\linewidth}{!}{
\begin{tabular}{llcccc}
\toprule
\makebox[45mm][l]{\textbf{Methods}} & \makebox[22mm][l]{\textbf{Epochs$^{1}$}} & \makebox[24mm][c]{\textbf{DOTA-v1.0}} & \makebox[24mm][c]{\textbf{DOTA-v1.5}} & \makebox[24mm][c]{\textbf{DOTA-v2.0}} & \makebox[24mm][c]{\textbf{RSAR}}\\
\hline
\multicolumn{6}{l}{$\blacktriangledown$ \textit{RBox-supervised OOD}} \\ \hline
RetinaNet{\tiny ICCV'17} \cite{lin2017focal} & $12$ & 68.69 & 60.57        & 47.00 &  57.67  \\
GWD{\tiny GWD'21} \cite{yang2021rethinking} & $12$ & 71.66 & 63.27        & 48.87 & 57.80 \\
FCOS{\tiny ICCV'19} \cite{tian2019fcos} & $12$ & 72.44 & 64.53        & 51.77    &  \textbf{66.66} \\
S$^2$A-Net{\tiny AAAI'22} \cite{han2022align} & $12$ & \textbf{75.81} & \textbf{66.53} & \textbf{52.39} & 66.47 \\
\hline
\multicolumn{6}{l}{$\blacktriangledown$ \textit{HBox-supervised OOD}} \\ \hline
H2RBox{\tiny ICLR'23} \cite{yang2023h2rbox} & $12$ & 70.05 & 61.70        & 48.68    &  \underline{49.92}    \\
H2RBox-v2{\tiny NIPS'24} \cite{yu2023h2rboxv2} & $12$ & \underline{72.31} & \underline{64.76} & \underline{50.33} &  \textbf{65.16} \\
AFWS{\tiny TGRS'24} \cite{lu2024afws} & $12$ & \textbf{72.55} & \textbf{65.92} & \textbf{51.73} & - \\
\hline
\multicolumn{6}{l}{$\blacktriangledown$ \textit{Point-supervised OOD}} \\ \hline
PointOBB{\tiny CVPR'24} \cite{luo2024pointobb} & $24^*+12$ & 30.08 & 10.66        & 5.53     &  13.80    \\
Point2RBox+SK$^\dagger${\tiny CVPR'24} \cite{yu2024point2rbox} & $12^*$ & 40.27 & 30.51        & 23.43    &  \underline{27.81}    \\
PointOBB-v3{\tiny IJCV'25} \cite{zhang2025pointobbv3} & $24^*$ & 41.20 & 31.25 & 22.82 & 15.84 \\
PointOBB-v3{\tiny IJCV'25} \cite{zhang2025pointobbv3} & $24^*+12$ & \underline{49.24} & \underline{33.79} & \underline{23.52} & 22.60 \\
\midrule
PointOBB-v2{\tiny ICLR'25} \cite{ren2024pointobbv2} & $6+12$ & 41.68 & 30.59        & 20.64    &  18.99   \\
PMS-SAM-RSD{\tiny EL'25}\cite{lu2025semantic} & 12 & 46.00 & - & - & - \\
\rowcolor{gray!20} 
SSP{\tiny Ours} & $6+12$ & \textbf{50.81} & \textbf{35.93} & \textbf{29.02} & \textbf{31.16} \\
\bottomrule
\specialrule{0pt}{2pt}{0pt}
\multicolumn{6}{l}{$^\dagger$Using strong priors. 
Point2RBox+SK: One-shot sketches for each class; PMS-SAM-RSD: SAM.} \\
\multicolumn{6}{l}{$^1$$^*$indicates data passes through teacher\&student two models  thereby doubling the equivalent epochs; $+$ denotes two-stage training.} \\
\bottomrule
\end{tabular}}
\caption{Accuracy (AP$_{50}$) comparisons on the DOTA-v1.0/1.5/2.0 and RSAR datasets.}
\label{tab:exp_other}
\end{table*}

\subsection{Main results on DOTA-v1.0}
Table \ref{tab:exp_dota} provides a detailed class-by-class comparison with various state-of-the-art (SOTA) methods on the most classic dataset DOTA-v1.0 in the field. Under the point supervision track, we further subdivide it according to the model paradigm (teacher-student vs. simple model). 
In terms of the overall metric mAP, SSP achieves significant improvement from 41.68 (of the baseline method PointOBB-v2) to 48.51, and outperforms the SAM-based PMS-SAM-RSD method (\textbf{48.41} vs. 46.00). Furthermore, even when directly competing in performance with methods based on the teacher-student paradigm, SSP still holds its own. It first markedly outperforms PointOBB, Point2RBox, and PointOBB-v3 (\textbf{48.41} vs. 30.08 vs. 34.07 vs. 41.29), and achieves performance comparable to Point2RBox-v2 (\textbf{48.41} vs. 51.00). Notably, these competing methods generally require far longer training times and much higher GPU memory consumption (see Table \ref{tab:exp_training_cost} for details); PointOBB-v3, in particular, requires an extensive 24 training epochs. Additionally, the performance can be further improved by using a more powerful detector, e.g., 50.00\% mAP with ORCNN detector and 50.81\% mAP with ReDet detector.
Notably, our method outperforms the baseline PointOBB-v2 significantly across diverse scene categories, particularly for \textit{ground-track-field (GTF)} (\textbf{42.0} vs. 36.2), \textit{basketball-court (BC)} (\textbf{68.3} vs. 62.2), \textit{soccer-ball-field (SBF)} (\textbf{33.3} vs. 12.1), and \textit{swimming-pool (SP)} (\textbf{55.9} vs. 43.7). This improvement stems from integrating dense arrangement priors via spatial partitioning and leveraging similar appearance features through region growing—mechanisms particularly effective for the aforementioned categories. Even \textit{harbors (HB)}, typical large-aspect-ratio objects, demonstrate substantial gains (\textbf{30.4} vs. 8.1). However, \textit{bridge (BR)} detection performance remains poor, on par with PointOBB-v2.
Dataset analysis reveals two key insights: 1) Harbors inhabit dense scenes, while bridges appear in sparse environments; 2) Harbors are surrounded by oceans with distinct foreground-background contrasts, whereas bridges connect to roads, often confused with road segments. These characteristics hinder bridges from benefiting from spatial partitioning and region growing, leading to subpar performance. These findings provide us with deeper understanding of our method.

\subsection{More results on other datasets}
Table \ref{tab:exp_other} provides an overall performance comparison with various state-of-the-art methods on other common datasets. On the more challenging DOTA-v1.5 and DOTA-v2.0 datasets, which consist of more complex scenes and object layouts, our method SSP still has a significant advantage over PointOBB-v2 (\textbf{33.52} v.s. 30.59, \textbf{25.36} v.s. 20.64). On the synthetic aperture radar (SAR) image dataset RSAR, our model has achieved a remarkable improvement (\textbf{31.16} v.s. 18.99). This may be because the SAR data has characteristics that make the region growing algorithm, a key component of the model, perform better and achieve good results. In summary, our method has achieved consistent performance improvements across multiple different datasets, further validating the robustness of our approach.

\begin{table}
\centering
\resizebox{\linewidth}{!}{
\begin{tabular}{lcccc}
\toprule
\textbf{Methods} & \textbf{Epochs} & \textbf{Train Time(Hours)} & \textbf{Memory(GB)} & \textbf{mAP}\\
\midrule
Point2RBox{\tiny CVPR'24}\cite{yu2024point2rbox} & $12^*$ & 9.21 & 13.17 & 40.27\\
Point2RBox-v2{\tiny CVPR'25}\cite{yu2025point2rbox-v2} & $12^*$ & 10.36 & 9.64 & 51.00 \\
PointOBB{\tiny CVPR'24}\cite{luo2024pointobb} & $24^*$ & 22.28 & $>24$ & 30.08\\
PointOBB-v3{\tiny IJCV'25}\cite{zhang2025pointobbv3} & $24^*$ & 24.22 & 20.23 & 41.29\\
\midrule
PointOBB-v2{\tiny ICLR'25}\cite{ren2024pointobbv2} & \underline{6} & \textbf{1.43} & \underline{5.99} & 41.68\\
PMS-SAM-RSD{\tiny EL'25}\cite{lu2025semantic}
 & 12 & - & - & 46.00 \\ \rowcolor{gray!20} 
SSP{\tiny Ours} & \textbf{6} & \underline{1.91} & \textbf{5.34} & 50.81\\
\bottomrule
\end{tabular}}
\caption{Train cost of different methods under 2 NVIDIA 3090 GPUs.}
\label{tab:exp_training_cost}
\end{table}

\subsection{Evaluation about training cost} 
Table \ref{tab:exp_training_cost} presents the evaluation results regarding the training cost of different methods. We strictly adhere to the experimental settings consistent with those in the same experiment of PointOBB-v2. 
Teacher-student models generally have high resource consumption. Among them, the Point2Rbox series consumes about 10 hours and 10 GB of GPU memory, while the PointOBB series even requires approximately 24 hours and 24GB of GPU memory. As a simple model like PointOBB-v2, SSP also only consumes less than 6GB of GPU memory and takes merely 2 hours to complete training, thus achieving truly efficient training.

\subsection{Ablation Studies}
In order to explore the importance of various design elements in model training, we conducted extensive ablation experiments on the DOTA-v1.0 dataset, and report the performance of pseudo-labels on the training set and the performance of the downstream detector on the test set. For the performance of pseudo-labels, in addition to the mAP metric, we also provided the mIoU metric as additional reference. This follows the experiment protocol in previous works \cite{luo2024pointobb,ren2024pointobbv2}. Since the pseudo-labels are actually in strict match with the true annotations, the mIoU can more directly reflect the quality of the labels.
Note that during selection of the best model setting, we primarily refer to pseudo-label metrics for convenience. However, empirical results show that neither the mAP nor the mIoU of pseudo-labels can absolutely reliably reflect the downstream detection performance, so the mAP of detection is also reported in the paper.

\subsubsection{Different sample assignment strategy}
We conducted an ablation study to explore the impact of the sample assignment strategy on the detection performance, as shown in Table \ref{tab:exp_sample_assignment_strategy}. \textit{Pos-radius} means regarding region near the object center as positive samples; \textit{Neg-radius} refers to treating areas outside the dynamic radius estimated by the layout as negative samples; \textit{Pos-growing} denotes taking the instance mask obtained through spatial partitioning with region growing as positive samples; \textit{Neg-partition} signifies regarding spatial partitioning boundaries as negative samples, as shown in Fig. \ref{fig:assignment}. Results show that relying solely on the original pos-radius strategy led to extremely limited performance (12.88 on mAP@DET). However, by introducing neg-radius, performance was significantly enhanced (\textbf{20.74} v.s. 12.88 mAP@DET) due to the reduction of numerous erroneous negative samples. Upon introducing pos-growing, more refined object shapes were obtained during sample allocation, resulting in a improvement in performance (\textbf{43.05} v.s. 20.74 on mAP@DET). Finally, with the addition of neg-partition, adjacent objects became easier to distinguish, yielding a modest further enhancement (\textbf{48.41} v.s. 43.05 on mAP@DET).

\begin{table}
\centering
\resizebox{\linewidth}{!}{
\begin{tabular}{ccccccc}
\toprule
\multirow{2}{*}{\textbf{Pos-radius}} & \multirow{2}{*}{\textbf{Neg-radius}} & \multirow{2}{*}{\textbf{Pos-growing}} & \multirow{2}{*}{\textbf{Neg-partition}} & \multicolumn{2}{c}{\textbf{PSE}} & \textbf{DET} \\
& & & & mIoU & mAP & mAP \\
\midrule
\checkmark & & & & 36.11 & 12.40 & 12.88 \\
\checkmark & \checkmark & & & 38.79 & 16.89 & 20.74\\
\checkmark & \checkmark & \checkmark & & \underline{47.13} & \underline{32.74} & \underline{43.05}\\
\rowcolor{gray!20} 
\checkmark & \checkmark & \checkmark & \checkmark & \textbf{49.51} & \textbf{36.80} & \textbf{48.41}\\ 
\bottomrule
\end{tabular}}
\caption{Ablation study of sample assignment.}
\label{tab:exp_sample_assignment_strategy}
\end{table}

\subsubsection{Spatial partition with different semantic strategy}
We conducted an ablation study to explore the impact of the semantic spatial partitioning strategy on the detection performance, as shown in Table \ref{tab:exp_semantic_spatial_partition.}. 
\textit{Pos-center} denotes just selecting central points as positive samples for region growing; \textit{Neg-boundary} refers to introducing spatial partitioning boundaries as negative samples for region growing; \textit{Pos/Neg-semantic} utilizes semantic scores to additionally pre-assign a portion of positive and negative samples. The results show that spatial partitioning boundaries can significantly improve the instance extraction effect, thereby enhancing the detection performance (\textbf{45.43} v.s. 37.57 on mAP@DET). On this basis, pre-assigning positive and negative samples by leveraging semantic scores can further improve detection performance, albeit relatively slightly (\textbf{48.41} v.s. 45.43 on mAP@DET). This may be because partitioning boundaries have already substantially reduced the scope of region growing, and its role heavily overlaps with that of the semantic score.

\begin{table}
\centering
\resizebox{\linewidth}{!}{
\begin{tabular}{cccccc}
\toprule
\multirow{2}{*}{\textbf{Pos-center}} & \multirow{2}{*}{\textbf{Neg-boundary}} & \multirow{2}{*}{\textbf{Pos/Neg-semantic}} & \multicolumn{2}{c}{\textbf{PSE}} & \textbf{DET} \\
& & & mIoU & mAP & mAP \\
\midrule
\checkmark & & & 43.91 & 27.39 & 37.57\\
\checkmark & \checkmark & & \underline{48.43} & \underline{33.20} & \underline{45.43}\\
\rowcolor{gray!20} 
\checkmark & \checkmark & \checkmark & \textbf{49.51} & \textbf{36.80} & \textbf{48.41}\\ 
\bottomrule
\end{tabular}}
\caption{Ablation study of semantic spatial partition.}
\label{tab:exp_semantic_spatial_partition.}
\end{table}

\subsubsection{Instance mask generation from different sources}
We conducted an ablation study to explore the impact of the source of instance masks on the detection performance, as shown in Table \ref{tab:exp_instance_mask_source}. Utilizing the spatial partitioning with region growing algorithms, we carried out instance extraction on the \textit{original image}, the \textit{merged semantic map}, and the \textit{decoupled semantic map} respectively, and then converted them into bounding boxes as pseudo labels. The results show that, compared with the original image, directly performing operations on the aggregated semantic map does not bring improvements (30.81 v.s. 3\textbf{4.66} on mAP@DET), but semantic decoupling can release the potential of the semantic map and yield significant improvements (\textbf{48.41} v.s. 30.81 on mAP@DET).

\subsubsection{Instance box extraction with different methods}
We conducted an ablation study to explore the impact of the instance bounding box extraction method on the detection performance, as shown in Table \ref{tab:exp_instance_box_extraction}. The most basic approach is the \textit{binarization} method, which binarizes the semantic map based on a fixed threshold to obtain instance masks, and further converts them into boxes as pseudo-labels. Due to its inability to effectively overcome interference among multiple instances and the impact of noise, this method exhibits the poorest performance (37.64 on mAP@DET). Employing the \textit{oriented-search} method used in PointOBB-v2, which involves estimating the direction and then searching for the object boundaries in that direction, can achieve a significant improvement (\textbf{42.80} vs. 37.64 on mAP@DET). Furthermore, by adopting the \textit{spatial partition} with region growing proposed in this paper, the results can be further enhanced (\textbf{48.41} v.s. 42.80 mAP@DET).

\begin{table*}[tb!]
\centering
\begin{minipage}[t]{0.31\linewidth}
\centering
\resizebox{1\linewidth}{!}{
\begin{tabular}{cccc}
\toprule
\multirow{2}{*}{\textbf{Instance Source}} & \multicolumn{2}{c}{\textbf{PSE}} & \textbf{DET} \\
& mIoU & mAP & mAP \\
\midrule
raw-image & 41.68 & \underline{24.43}  & \underline{34.66}\\
merged-semantic & \underline{42.10} & 23.69  & 30.81\\
\rowcolor{gray!20} 
decoupled-semantic & \textbf{49.51} & \textbf{36.80} & \textbf{48.41}\\ 
\bottomrule
\end{tabular}}
\caption{Ablation study of instance mask source.}
\label{tab:exp_instance_mask_source}
\end{minipage}
\quad
\begin{minipage}[t]{0.31\linewidth}
\centering
\resizebox{1\linewidth}{!}{
\begin{tabular}{cccc}
\toprule
\multirow{2}{*}{\textbf{Instance extraction}} & \multicolumn{2}{c}{\textbf{PSE}} & \textbf{DET} \\
& mIoU & mAP & mAP \\
\midrule
binarization & 43.93 & 27.31 & 37.64 \\
oriented-search & \underline{46.31} & \underline{30.27} & \underline{42.80}\\ 
\rowcolor{gray!20} 
spatial-partition & \textbf{49.51} & \textbf{36.80} & \textbf{48.41}\\ 
\bottomrule
\end{tabular}}
\caption{Ablation study of instance box extraction.}
\label{tab:exp_instance_box_extraction}
\end{minipage}
\quad
\begin{minipage}[t]{0.30\linewidth}
\centering
\resizebox{1\linewidth}{!}{
\begin{tabular}{cccc}
\toprule
\multirow{2}{*}{\textbf{Annotation offsets}} & \multicolumn{2}{c}{\textbf{PSE}} & \textbf{DET} \\
& mIoU & mAP & mAP \\
\midrule
10\% & \textbf{48.36} & \textbf{35.21} & \textbf{45.79}\\
20\% & \underline{45.43} & \underline{30.94} & \underline{42.20}\\
30\% & 42.05 & 19.92 &  38.79\\
\bottomrule
\end{tabular}}
\caption{Ablation study of annotation offsets.}
\label{tab:exp_point_annotation_offsets}
\end{minipage}
\end{table*}

\begin{figure*}[!t]
\centering
\includegraphics[width=0.85\linewidth]{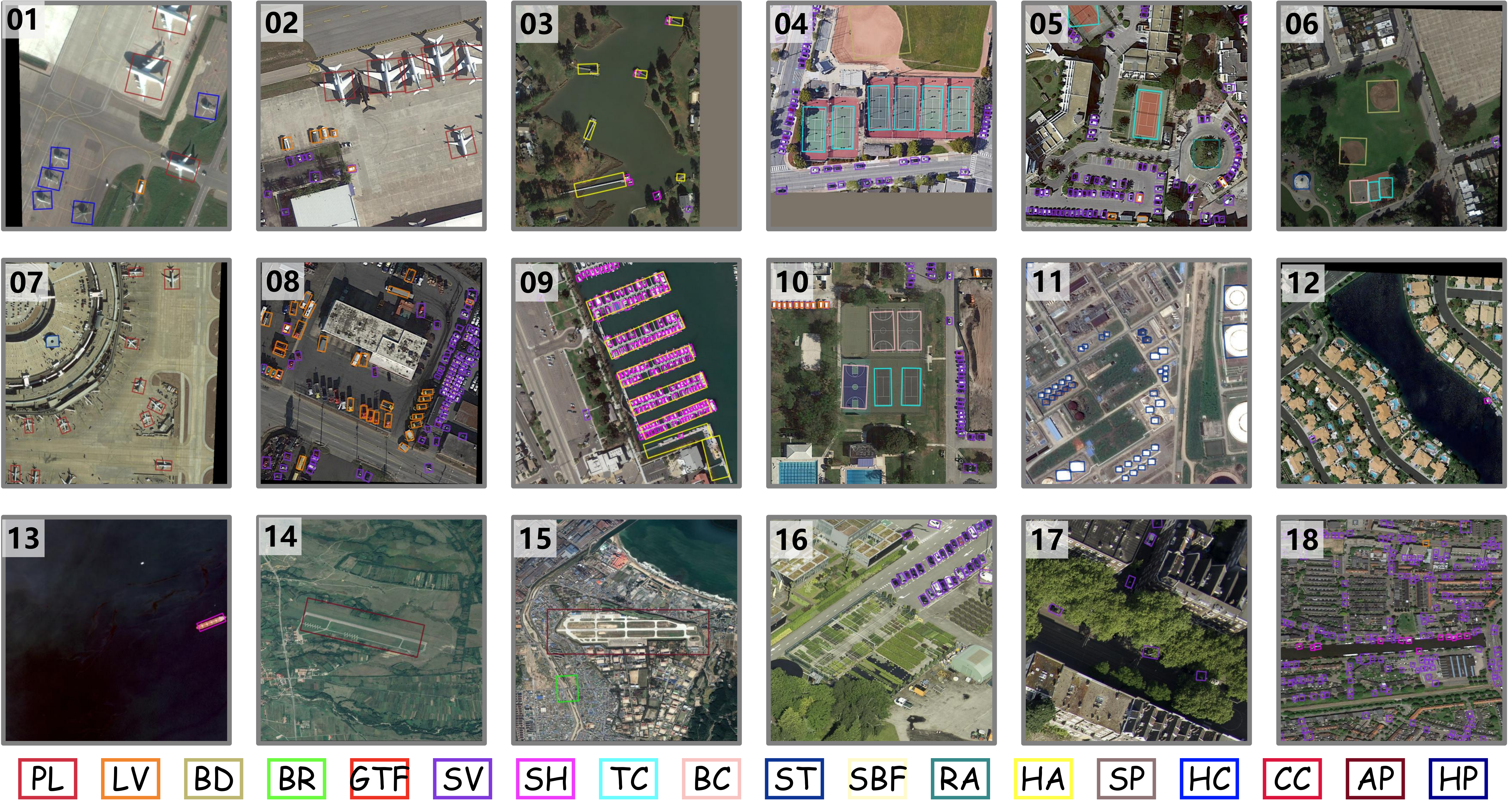}
\caption{Visualized detection results on DOTA datasets. Image \#01$\sim$\#12 are from DOTA-v1.0, and the other are from DOTA-v1.5/v2.0.}
\label{fig:vis_dota}
\end{figure*}

\begin{figure*}[!t]
\centering
\includegraphics[width=0.85\linewidth]{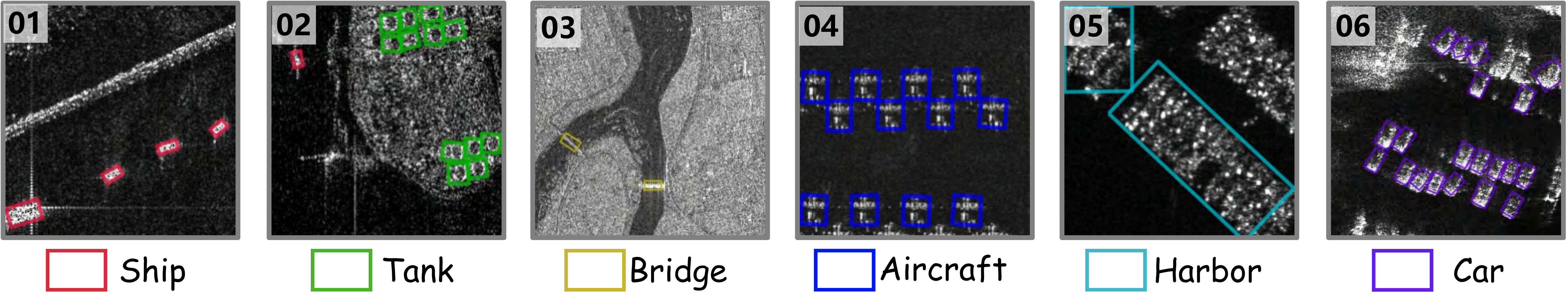}
\caption{Visualized detection results on RSAR dataset.}
\label{fig:vis_rsar}
\end{figure*}

\begin{figure*}[!t]
\centering
\includegraphics[width=0.85\linewidth]{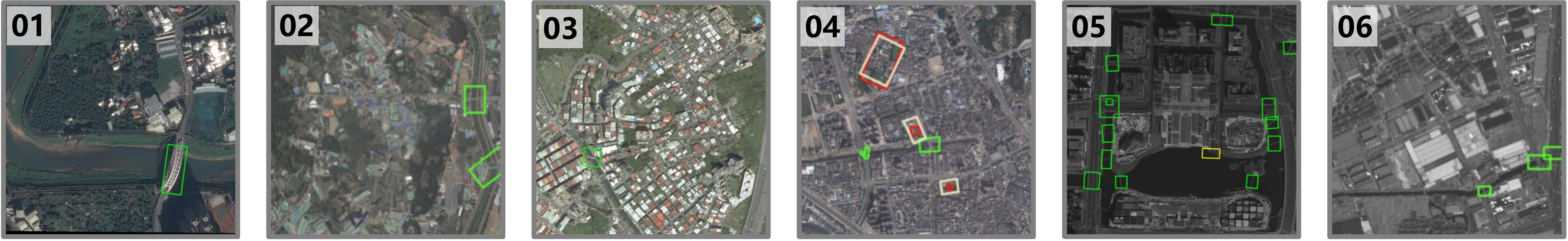}
\caption{Bad cases about bridge on DOTA dataset.}
\label{fig:vis_bridge}
\end{figure*}

\subsubsection{Point annotation of different offsets}
We conducted an ablation study to explore the impact of annotation point offset on the detection performance, as is shown in Table \ref{tab:exp_point_annotation_offsets}. Following the implementation in previous works \cite{yu2023h2rboxv2,yu2024point2rbox,ren2024pointobbv2}, we randomly offset the annotation points within the range from \textit{10\%} to \textit{30\%} of the object's width and height. The result shows that a slight offset of 10\% has a negligible impact on the detection performance (\textbf{45.79} v.s. 48.41 on mAP@DET). However, when the offset reaches 30\%, the accuracy significantly decreases (36.05 v.s. \textbf{48.41} on mAP@DET), but it is still higher than some methods without offset, such as PointOBB (\textbf{38.79} v.s. 30.08) and Point2RBox (\textbf{38.79} v.s. 34.07). This indicates that our method possesses a certain level of robustness, such that annotation points only need to be roughly located near the object center rather than strictly at the geometric center.

\subsection{Visualized Analysis}
We provide some visualized detection results on the DOTA-v1.0/v1.5/v2.0 dataset as shown in Fig. \ref{fig:vis_dota} and RSAR dataset as shown in Fig. \ref{fig:vis_rsar}. While the model is trained under point supervision, it still demonstrates satisfactory detection performance in these typical scenes, with only slight deviations in scale and orientation. 
\abcmark{Notably, even for dense scenes with nested objects (e.g., ships berthed in harbors in Fig. \ref{fig:vis_dota}(image \#09)), our class-decoupled partition and category compatibility design effectively eliminate mutual detection interference between paired categories.}
\abcmark{We also conduct bad case analysis for bridges, whose mAP is significantly lower than other categories. As shown in Fig. \ref{fig:vis_bridge}, bridges exhibit diverse shapes and significant aspect ratio variations. Missing objects are prevalent, and even detected instances are often classified as false positives due to extremely low IoU with ground truth. This is an issue commonly encountered in current point-supervised methods, warranting urgent resolution in future research.}

\section{Conclusion}
\label{sec:conclusion}
In this paper, we explored the task of point-supervised oriented object detection within the simple-model paradigm. We identified two major issues in the previous state-of-the-art methods, i.e., 1) inadequate sample assignment for mask learning, and 2) unstable instance discrimination for box extraction. To address these problems, we proposed a training-efficient method named SSP, with its core focusing on spatial partitioning with region growing. 
It introduces strong prior knowledge through rule-based sample assignment and then filters label noise via model learning to obtain reliable boxes. 
\abcmark{SSP fully retains the native training efficiency of simple-model paradigm while significantly improving detection performance.}
It achieves +6.73\% mAP improvement compared with the baseline method PointOBB-v2 on the DOTA-v1.0 dataset, while requiring only 2 hours of training time and 6 GB of GPU memory. Experiments on more datasets and detectors further verified the effectiveness of this method. 

\abcmark{
While our SSP framework achieves a favorable efficiency-performance trade-off for point-supervised oriented object detection in typical scenarios, it has inherent limitations. Our method is built on two widely-validated priors for most orthographic scenes: minimal intra-class object overlap, and distinguishable foreground-background appearance. We design a class-decoupled partition strategy to eliminate possible cross-class overlap interference, and perform region growing based on model-learned semantic features to suppress imaging noise interference. Despite these designs, performance may degrade in extreme prior-violating scenarios (e.g., dense intra-class overlap, background-homogeneous bridges). Additionally, pre-defined category compatibility pairs need extra domain priors for unseen dataset generalization, and extreme noise/ultra-low resolution may still cause performance fluctuations. These common challenges for rule-based weakly supervised pseudo-labeling also inform our future work.
}

\section*{Acknowledgements}
This work is supported by National Key R\&D Program of China (2022YFD2001601), National Natural Science Foundation of China (62372433) and Zhejiang Provincial Natural Science Foundation of China (LQN25F020015).


\bibliographystyle{elsarticle-num} 
\bibliography{reference}

\end{document}